\documentclass[12pt]{article}
\usepackage[ruled,vlined]{algorithm2e}
\usepackage{amsmath,amssymb,mathtools}
\usepackage{graphicx,subcaption}
\usepackage{enumitem}
\usepackage{natbib}
\usepackage{url} 

\newcommand{\blind}{0}

\begin{document}

\def\spacingset#1{\renewcommand{\baselinestretch}%
{#1}\small\normalsize} \spacingset{1}

%%%%%%%%%%%%%%%%%%%%%%%%%%%%%%%%%%%%%%%%%%%%%%%%%%%%%%%%%%%%%%%%%%%%%%%%%%%%%%

\if0\blind
{
  \title{\bf Capturing patterns of variation unique to a specific dataset}
  \author{Robin Tu \\
    Department of Statistics, University of Illinois at Urbana-Champaign\\
    Alexander H. Foss \\
    Statistical Sciences, Sandia National Laboratories\\
    Sihai D. Zhao\\
    Department of Statistics, University of Illinois at Urbana-Champaign\\}
  \maketitle
} \fi

\if1\blind
{
  \bigskip
  \bigskip
  \bigskip
  \begin{center}
    {\LARGE\bf Title}
\end{center}
  \medskip
} \fi

\bigskip
\begin{abstract}
Capturing patterns of variation present in a dataset is important in exploratory data analysis and unsupervised learning. Contrastive dimension reduction methods, such as contrastive principal component analysis (cPCA), find patterns unique to a target dataset of interest by contrasting with a carefully chosen background dataset representing unwanted or uninteresting variation. However, such methods typically require a tuning parameter that governs the level of contrast, and it is unclear how to choose this parameter objectively. Furthermore, it is frequently of interest to contrast against multiple backgrounds, which is difficult to accomplish with existing methods. We propose unique component analysis (UCA), a tuning-free method that identifies low-dimensional representations of a target dataset relative to one or more comparison datasets. It is computationally efficient even with large numbers of features. We show in several experiments that UCA with a single background dataset achieves similar results compared to cPCA with various tuning parameters, and that UCA with multiple individual background datasets is superior to both cPCA with any single background data and cPCA with a pooled background dataset.
\end{abstract}

\noindent%
\vfill

\newpage
\spacingset{1.5} % DON'T change the spacing!
\section{Introduction}
\label{sec:intro}

Capturing patterns of variation present in a dataset is an important step in exploratory data analysis and unsupervised learning. Popular methods include principal component analysis (PCA) by \cite{pca}, nonnegative matrix factorization by \cite{Lee1999}, projection pursuit by \cite{pp}, and independent component analysis by \cite{ica}. Frequently, however, many of the identified patterns may actually arise from systematic or technical variation, for example batch effects, that are not of substantive interest.

New approaches are necessary for capturing meaningful patterns of variation. A popular approach is to contrast a target dataset of interest with a carefully chosen background dataset that represents unwanted or uninteresting variation. Patterns of variation unique to the target, and not present in the background, are more likely to be substantively meaningful.
For example, in Section \ref{sec:mouse}, we analyze proteomics data from of normal and trisomic mice that have undergone Context Shock treatment and/or drug therapy. One goal is to identify patterns in normal mice relative to trisomic mice. However, the dominant modes of variation in the dataset arise from the Context Shock treatment and the trisomic gene, which are not of interest. Therefore, we contrast the data with a background dataset of proteomics data from Context Shock-treated and drug-treated trisomic mice. As a result, the remaining patterns of variation unique to the target dataset reveal features specific to normal mice.
%For example, in Section \ref{sec:faces}, we analyze images of the faces of actors expressing different emotions. Our goal is to identify features uniquely associated with expression of fear (labeled as afraid). However, the dominant modes of variation in the dataset correspond to general variation in facial features that are not meaningful for the emotion of interest. Therefore, we contrast the data with a background dataset of images of other emotions that are not of interest. As a result, the remaining patterns of variation unique to the target dataset reveal features specific to afraid.

This approach was first introduced by \cite{Abid}, who proposed contrastive principal components analysis (cPCA). While standard PCA identifies directions of variation that explain the most variation in the target dataset, cPCA seeks directions that explain more variation in the target than in the background. The most important patterns are those corresponding to the largest gap between the two datasets. \cite{Salloum} later introduced cPCA++, which maximizes the ratio, rather than the difference, of the variance explained by the target and background. \cite{Boileau} described sparse cPCA, which seeks maxmially contrastive patterns of variation that can be characterized using a parsimonious set of features. Other types of contrastive implementations include latent models by \cite{severson2019unsupervised} and autoencoders by \cite{cautoencoder}.

There are two main issues with existing contrastive methods. First, a major disadvantage is that they cannot accommodate multiple background datasets. Using multiple backgrounds allows researchers to better hone in on the unique variation of interest by removing multiple types or sources of unwanted variation.
%referencing section faces is awkward cause we never talk about mice. should we do mice too?
For example, in the emotion analysis in Section \ref{sec:faces}, where we seek to uncover facial features characterizing the expression of ``afraid'', the dataset we use also contains background images from six other emotions. If we can simultaneously contrast the target data with multiple other background emotions, we will be able to identify more refined and distinctive patterns of variation characterizing ``afraid''. Naively applying existing methods by pooling the different emotions into a single background dataset is suboptimal, as variation in the pooled data may not be representative of variation in any of the individual datasets.

The second disadvantage of existing contrastive methods is that they typically require one or more tuning parameters. For example, cPCA requires the user to specify how much to penalize patterns that explain a large amount of variation in the background data. It is not clear in general how to choose these tuning parameters objectively.

We propose Unique Component Analysis (UCA), which addresses both of these issues. 
Not only can UCA contrast a target dataset against multiple backgrounds, but it also does not require any tuning parameters. UCA finds directions of variation that maximizes the explained variation in the target under a constraint on the amount of variation they can explain in each of the backgrounds. With a single background, UCA is equivalent to cPCA but with an automatically selected tuning parameter. We show that UCA achieves similar results as cPCA and cPCA++ with a single background and that it can outperform them when using multiple backgrounds. We also develop computationally scalable algorithms for application to experiments with large numbers of measured features.

The remainder of this article is organized as follows. In section~\ref{sec:background}, we present a brief overview of cPCA and cPCA++. Next, in section~\ref{sec:method}, we formally define UCA and mathematically detail our tuning-free, multibackground solution. Further, we present a computational algorithm to run UCA without the formation of covariance matrices. In section~\ref{sec:results} we demonstrate UCA's superiority over cPCA and cPCA++ in two real world data examples involving mice protein expression and the Karolinska Directed Emotional Faces dataset. Additionally, we illustrate high-dimensional computational improvements by circumventing the formation of covariance matrices. Lastly, we conclude by reviewing the three main advantages UCA has over existing contrastive methods.

\subsection{Background\label{sec:background}}
\subsubsection{Contrastive principal component analysis }
We first briefly review cPCA \cite{Abid}. Let $A$ denoted the $p\times p$ sample covariance matrix constucted from target data $X_{n_x \times p}$ and $B$ denote the $p\times p$ sample background covariance constructed from background data $Y_{n_y \times p}$. For some parameter $\lambda$, cPCA seeks eigenvectors $\hat{v}_\lambda \in \mathbb{R}^p$ that account for large variation in the target data $Y$ and small variation in the background data $X$ such that $\|\hat{v}_\lambda\|_2 = 1$. Specifically the first eigenvector is
\begin{equation}
  \label{eq:cpca}
  \begin{aligned}
  \hat{v}_\lambda = \text{arg}\max_{v \in \mathbb{R}^p}{v^T\left(A - \lambda B\right)v} \\\mbox{ subject to } v^\top v = 1 
  \end{aligned}
\end{equation}
and, for example, the second eigenvector maximizes the quadratic form subject to be orthogonal to $\hat{v}_\lambda$.

For a given $\lambda$, $\hat{v}_\lambda$ can be interpreted as the direction that maximizes the variation in $A$ without explaining much variation in $B$. The tuning parameter $\lambda$ measures how much to penalize the background data covariance. When $\lambda = 0$, background variation is not considered, therefore cPCA reduces to PCA. As $\lambda$ increases, the relative background variation becomes more dominant, causing $\hat{v}_\lambda$ to focus on directions which minimize background variation rather than maximizing the target. For very large values of $\lambda$, $\hat{v}_\lambda$ is equivalent to PCA after projecting the target data onto the nullspace of the background data.  The authors of cPCA suggested that $\lambda$ be chosen using spectral clustering via a parameter sweep of logarithmically spaced candidate values.

\subsubsection{cPCA++}
%\textbf{Why does cPCA++ enforce $v^\top B v = 1$? We should explain.} It's not enforced, but 
%\textbf{How is this possible? Need to explain. Basically we need to explain how cPCA++ works before we explain UCA.}
To eliminate the tuning parameter $\lambda$, Salloum and Kuo  \cite{Salloum} proposed cPCA++, where they seek to maximize the ratio, rather than the difference, of the variance explained in the target to the variance explained in the background.
%motivated by image splicing localization, classifying whether a pixel belonged to the original image or an image splice. Using a generalized likelihood ratio test framework to test whether a pixel belonged to foreground or background, they arrived at a Rayleigh Quotient of target variation divided by background variation.
Using similar notation as above where $A$ denotes the target sample covariance matrix and $B$ denotes the background sample covariance matrix, the first eigenvector of cPCA++ seeks to solve the generalized eigenvalue problem
\[\hat{v} = \text{arg}\max_{v\in \mathbb{R}^p} \frac{v^T A v}{v^T B v}.\]
To compare to cPCA \eqref{eq:cpca}, we can also write the cPCA++ objective function in its primal form \cite{ghojogh2019eigenvalue}:
\begin{equation}
  \label{eq:cpca++}
  \hat{v} = \text{arg}\max_{v \in \mathbb{R}^p}{v^T A v}  \mbox{ subject to } v^T B v = 1.
\end{equation}
While maximizing the relative variability between $A$ and $B$ is tuning parameter free, the solution involves a matrix inversion, which may not be feasible in high-dimensional applications like genomics where the inverse may not exist.  Furthermore, there is no clear path to extend this problem to multiple backgrounds.
\section{Material and Methods\label{sec:method}}
\subsection{Unique Component Analysis}
We introduce the Unique Component Analysis (UCA) framework, which combines ideas from both cPCA and cPCA++. First, we provide a reinterpretation of standard PCA. Applying PCA to a target dataset with covariance matrix $A$ can be viewed as finding $v$ that maximize the variance explained in the target, subject to explaining unit variance in a ``background'' dataset with covariance matrix $I$, where $I$ is a $p \times p$ identity matrix. It is natural to use this white noise as a baseline because it contains no patterns of variation, as all features are uncorrelated. Thus any informative eigenvector should explain more variance in the target than in a white noise background (provided that features in the target dataset are scaled to unit variance).

This interpretation reveals some issues with the formulations of both cPCA and cPCA++. From \eqref{eq:cpca}, we can see that cPCA requires that its eigenvectors explain unit variance in a white noise background, but somewhat arbitrarily combines both $A$ and $B$ into a target matrix. Conversely, \vphantom{\eqref{eq:cpca++}} shows that cPCA++ uses $A$ as the target but no longer requires its eigenvectors to explain unit variance in a white noise background. As a result, its eigenvectors are no longer comparable to those found by PCA and may actually explain less variance in the target than PCA eigenvectors.

To resolve these issues, we propose UCA, which joins the constraint of $v^T  v = 1$ from PCA and cPCA with the constraint $v^T Bv = 1$ from cPCA++. For a single background dataset, we propose to obtain the most important directions of variation by standardizing the features of the target and background to unit variance and then solving

\begin{equation}
  \label{eq:1}
  \begin{aligned}
  \max_{v\in \mathbb{R}^p}{\frac{v^TAv}{v^T v}} \text{ subject to }\;\; \frac{v^TBv}{v^T v} \leq 1.
  \end{aligned}
\end{equation}
To make our procedure comparable to PCA, we constrain the eigenvectors to explain exactly unit variance in the white noise background. In addition, we require them to explain no more than unit variance in the background dataset. Notably, \eqref{eq:1} does not require any tuning parameters.

Instead of directly solving \eqref{eq:1}, we instead solve the dual problem
\begin{align}
  \max_{\lambda \geq 0}{g(\lambda)} &= \max_{\lambda \geq 0}{\max_{v\in \mathbb{R}^P}{\mathcal{L}}}\left(v,\lambda\right) \nonumber\\
                                    &= \max_{\lambda \geq 0}{\max_{v\in \mathbb{R}^P}{\left( \frac{v^TAv}{v^T v} - \lambda\left(\frac{v^TBv}{v^T v} - 1\right)\right)}} \nonumber\\
                                    &= \max_{\lambda \geq 0}{\max_{v\in \mathbb{R}^P}{\left(\frac{v^T\left(A - \lambda B\right)v}{v^T v} + \lambda\right)}}\nonumber\\
                                        &= \max_{\lambda \geq 0}{\left(\lambda_{\text{max}}\left(A - \lambda B\right) + \lambda\right)}, \label{eq:2}
\end{align}
where $\lambda$ is a Lagrange multiplier and $\lambda_{\max}(A - \lambda B)$ is the largest eigenvalue associated with $A - \lambda B$. This is convenient because dual functions are always concave \cite{boyd2004convex}. We employ L-BFGS-B by \cite{byrd1995limited} to find the optimal $\lambda$ and speed up convergence by calculating the gradient of $g(\lambda)$. If $\hat{v}_\lambda$ is the first eigenvector of $A - \lambda B$, normalized such $\hat{v}_{\lambda}^\top \hat{v}_{\lambda} = 1$, implicit differentiation can be used to show that
\[
  \frac{d g}{d \lambda} = - \hat{v}^T _{\lambda} B \hat{v}_{\lambda} + 1.
\]

Finally, let $\hat{\lambda}$ denote the solution to the dual problem \eqref{eq:2}. Then we take the corresponding eigenvectors of the matrix $A - \hat{\lambda} B$ to be the components of our UCA algorithm. In this sense, UCA can be thought of as an automatically tuned version of cPCA with a contrastive parameter equal to $\hat{\lambda}$.

A major advantage of our UCA formulation is that we can accommodate multiple backgrounds. Let the $A$ be the target $p \times p$ covariance matrix and $ B_1, \ldots, B_m$ now be the $m$ background $p \times p$ covariance matrices, constructed from an $n_y \times p$ dimensional target data matrix $Y$ and corresponding $(n_{x_1} \times p), \ldots, (n_{x_m}\times p)$ dimensional background data matrices $X_1, \ldots, X_m$. We scale all features in all datasets to unit variance.

For each background $j = 1, \ldots, m$, we add additional constraints $\frac{v^TB_jv}{v^Tv}\leq 1$ to our optimization problem. Specifically, for multiple backgrounds, the primal problem becomes
\begin{equation}
  \label{eq:4}
  \begin{aligned}
  \max_{v\in \mathbb{R}^p}{\frac{v^TAv}{v^T v}} \\ \text{ subject to }\; \; &\frac{v^TB_1 v}{v^T v} \leq 1, \ldots, \frac{v^T B_m v}{v^T v}\leq 1.
  \end{aligned}
\end{equation}
The corresponding dual function problem is

\begin{align}
    &\max_{\lambda_1,\ldots,\lambda_m \geq 0}{g(\lambda_1,\ldots,\lambda_m)} \nonumber \\ % &= \max_{\lambda_1,\ldots,\lambda_m \geq 0}{\max_{v\in \mathbb{R}^{P}}{\mathcal{L}\left(v, \lambda_1, \ldots, \lambda_m\right)}} \nonumber \\
  %&=\max_{\lambda_1,\ldots,\lambda_m \geq 0}{\max_{v\in \mathbb{R}^{P}}{\left(v^TAv - \lambda_1\left(v^TB_1v - 1\right)-\ldots -\lambda_m\left(v^TB_mv - 1\right)\right)}}\nonumber \\
   &=\max_{\lambda_1,\ldots,\lambda_m \geq 0}{\max_{v \in \mathbb{R}^P}{\left(\frac{v^T\left(A - \sum^{m}_{j = 1}{\lambda_j B_j}\right)v}{v^T v} + \sum^{m}_{j=1}{\lambda_j}\right)}}\nonumber \\
   &=\max_{\lambda_1,\ldots,\lambda_m \geq 0}{\left(\lambda_{\text{max}}\left(A - \sum^{m}_{j = 1}{\lambda_j B_j}\right) + \sum^{m}_{j=1}{\lambda_j}\right)}.\label{eq:6}
\end{align}
We use coordinate descent to solve for $\hat{\lambda}_1, \ldots, \hat{\lambda}_m$, the solution to the dual problem \eqref{eq:6}. Then we take the corresponding eigenvectors of the matrix  $A - \sum_j \hat{\lambda}_j B$ to be the components of our UCA algorithm.

\subsection{Extension to High Dimensional Data:}

% \textbf{First need to explain what the problem is when we have high-dimensional data. Actually first need to explain what high-dimensional data means.}
Under the high-dimensional situation, where the number of variables $p$ far exceeds the number of samples $n$, constructing the covariance matrix and using eigendecomposition to find the top eigenvectors becomes computationally expensive. To implement standard PCA, we can avoid creating and storing a large $p \times p$ covariance matrix by instead applying singular-value decomposition (SVD) to the data matrix.

Analogously, to extend UCA to high-dimensional data, we introduce the Product SVD method to exploit the structure of the contrastive problem so that we can use SVD and avoid constructing $p \times p$ matrices. For a single background, let the target data $Y$ and the background data $X$ be centered and scaled $n_y \times p$ and $n_x \times p$ matrices, respectively. If $A$ and $B$ are the corresponding covariance matrices, we can write $A - \lambda B$ as a product of a $p \times (n_y + n_x)$ dimensional left matrix, $L$, and a $(n_y + n_x) \times p$ dimensional right matrix, $R$, as seen in equation \ref{eq:7}:
\begin{align}
  C_\lambda &= A - \lambda B \nonumber\\
            &=\frac{1}{n_y}Y^TY -\frac{\lambda}{n_x} X^T X\nonumber \\
            &=  \underbrace{\left[ \frac{1}{\sqrt{n_{y}}}Y^T,  \frac{-\lambda}{\sqrt{n_{x}}} X^T\right]}_{L}\underbrace{\begin{bmatrix*} \frac{1}{\sqrt{n_{y}}}Y \\ \frac{1}{\sqrt{n_{x}}}X \\ \end{bmatrix*}}_{R} \label{eq:7}
\end{align}
With this formulation, we can follow the steps in \cite{Golub}, and find the singular values and vectors of $C_\lambda$ using a sequence of SVD and QR decompositions operating on the left and right matrices. Since singular values and eigenvalues coincide in square matrices, we use the singular vectors, $U$, to find the largest eigenvalues by sorting the diagonal of $ULRU^T$.  
We describe the Product SVD Method in algorithm \ref{algo:product-svd}, which can directly replace the more computationally expensive eigendecomposition in high dimensions.

\begin{algorithm}[ht]
    \caption{Product SVD Method to find largest eigenvalue of $C_\lambda$}
  \label{algo:product-svd}
  \SetAlgoLined
  \textbf{Input:} Centered background matrix $X$, centered target matrix $Y$, and $\lambda$\;
  \nl Construct 
  $  L = \left[ \frac{1}{\sqrt{n_{y}}}Y^T, - \frac{\lambda}{\sqrt{n_{x}}} X^T\right],\;
  R = \begin{bmatrix*} \frac{1}{\sqrt{n_{y}}}Y \\ \frac{1}{\sqrt{n_{x}}}X \\ \end{bmatrix*} $\;
  \nl  Find the SVD of the right matrix, $R = U_R S_R V^T_R$ \;
  \nl  Find the QR decomposition of $LU_R$ into  $Q_{LU_R}R_{LU_R}$ \;
  \nl  Find the SVD of the product of $R_{LU_R}$ (step 3) and $S_R$ (step 2), $R_{LU_R}S_{R} = EDF^T$ \;
  \nl  The singular vectors of $C_\lambda$ are the product of $Q$ (step 3) and $E$ (step 4), $U_{C_\lambda} = Q_{LU_R}E$ \;
  \nl  Find the largest eigenvalues of $C_\lambda$ by sorting the diagonal of $D_{C_\lambda}$, where $D_{C_\lambda} = U_{C_\lambda} C_\lambda U_{C_\lambda}^T$ \;
  \textbf{Output:} $\lambda_{\text{max}}\left( C_\lambda \right)$ 
\end{algorithm}

Similarly, for multiple backgrounds, again let the $A$ be the target $p \times p$ covariance matrix and $ B_1, \ldots, B_m$ be the $m$ background $p \times p$ covariance matrices constructed from a $n_y \times p$ dimensional Y data matrix and corresponding $(n_{x_1} \times p), \ldots, (n_{x_m}\times p)$ dimensional $X_1, \ldots X_m$ background data matrices.

We can construct $C_{\boldsymbol{\lambda}} = A - \sum^{m}_{j=1}\lambda_jB_j$ analogously by appending the additional datasets to the left and right matrices:
\begin{align}
    C_{\boldsymbol{\lambda}}&= A - \sum^{m}_{j=1}\lambda_jB_j \nonumber \\
                                  &=\frac{1}{n_y}Y^{T}Y -\sum_{j=1}^{m}{\frac{\lambda_{j}}{n_{x_j}}X_{j}^TX_{j}} \nonumber\\
                                  &=  \underbrace{\left[\frac{1}{\sqrt{n_y}}Y^T, \frac{-\lambda_1}{\sqrt{n_{x_{1}}}} X^T_1, \ldots, \frac{-\lambda_m}{\sqrt{n_{x_{m}}}}X^T_m\right]}_{L} \underbrace{\begin{bmatrix} \frac{1}{\sqrt{n_{y}}}Y \\ \frac{1}{\sqrt{n_{x_{1}}}}X_1 \\ \vdots \\ \frac{1}{\sqrt{n_{x_{m}}}}X_m \end{bmatrix}}_{R} \label{eq:8}
\end{align}
We can apply this Product SVD method (algorithm \ref{algo:product-svd}) to solve the UCA dual problem.

The Product SVD method is advantageous in high-dimensions because it never explicitly operates on the entire $p \times p$ covariance matrix. 
Not only is our method more memory efficient, but our method is also computationally more efficient at finding the largest eigenvalue/eigenvector. By operating on the data matrix, our method scales with $n\times p$ bytes rather than $p^2$ bytes. Further, operating SVD and QR decomposition on either the left or right matrices is faster than directly operating eigendecomposition on the covariance matrix. In the single background scenario,  $\lambda$ only appears in $L$. Thus since, $R$ doesn't change, we can pre-compute the SVD of $R$, and only update $L$ when solving for $\lambda_{\text{max}}$.
Similarly, in the multi-background scenario, rather than reconstructing $A^{*}_{(j)} = A-\sum^{m}_{i\neq j}\lambda_i B_i$ for each $j$ in the original coordinate descent algorithm, this method allows us to only modify the $j$ element of the left data matrix, $L$.
The SVD of the right matrix, $R$, only needs to be computed once in our coordinate descent. Furthermore, at each step within the coordinate descent only step 3 and 4, the SVD and QR are computed, and are done on matrices with dimensions much smaller than $p \times p$.

All computations in this paper were done with R 3.6.3 \cite{baseR} on an AMD Ryzen 1700X 3.7 gHz processor and 64GB 3000 mhz DDR4 RAM, Fedora 33 system. 

\subsection{Algorithm and software implementation}
We have released a R implementation of UCA which is downloadable at \url{https://github.com/rtud2/Unique-Component-Analysis}. We have implemented both Product SVD on the data matrix and eigendecomposition on the contrastive covariance matrix and allow the background to take either a single background or a list of backgrounds. The GitHub repository also includes R markdown and datasets that reproduce most of the figures in this paper and in the Supplementary.

\subsection{Data Availability}
Datasets that have been used to evaluate UCA in this paper are publicly available from the authors of the original studies. The mouse proteomics data are available from the UC Irvine Machine Learning Repository \url{https://archive.ics.uci.edu/ml/datasets/Mice+Protein+Expression} and the Karolinska Directed Emotional Faces (KDEF) are available from \url{https://www.kdef.se/}.
\section{Results\label{sec:results}}
\subsection{\label{sec:mouse}Discovering subgroups in protein expression data}

\begin{figure*}[t!]
  \centering
  \includegraphics[width = 0.8\textwidth]{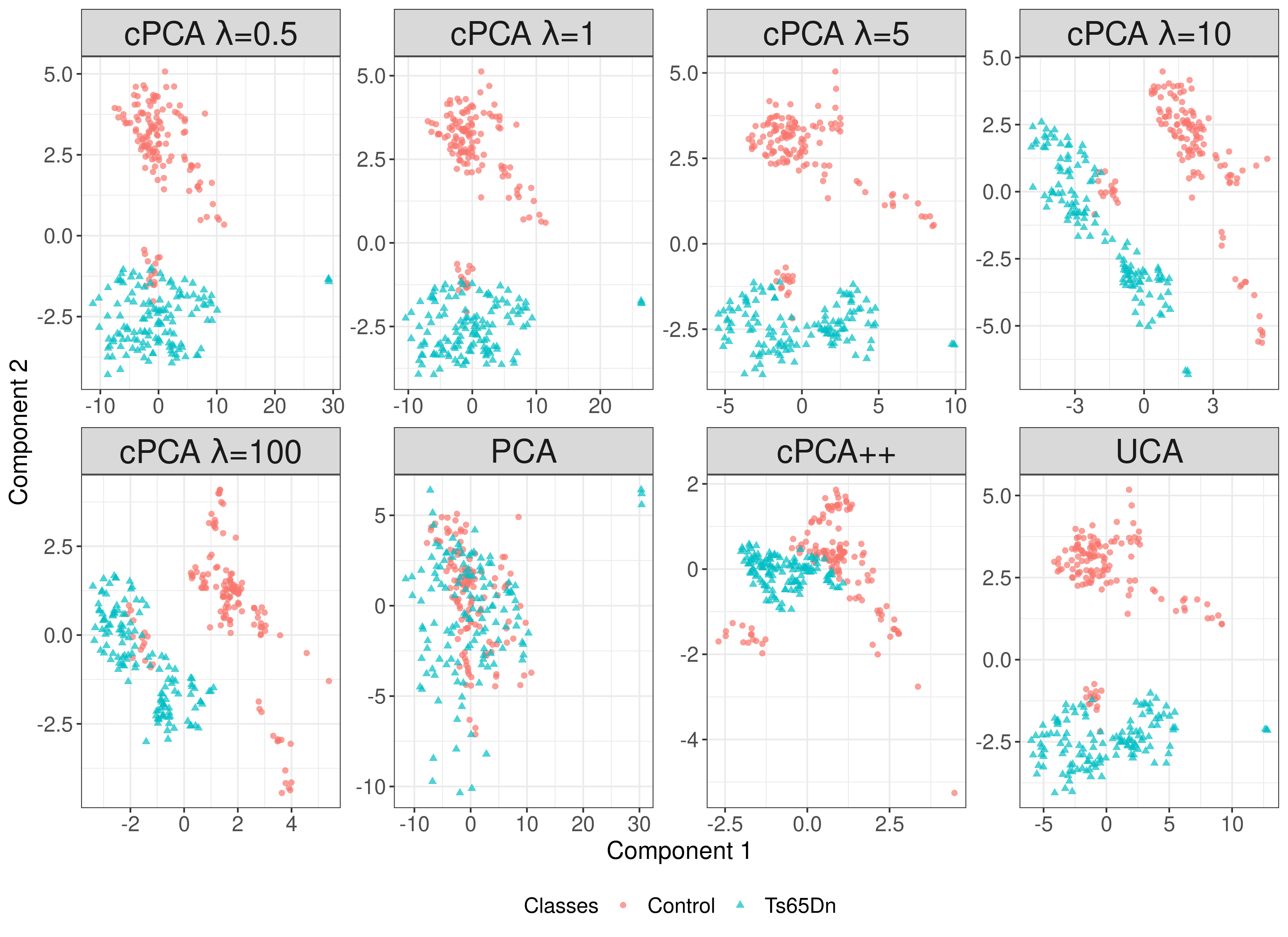}
  \caption{Mouse Protein Expression: Normal and trisomic (Ts65Dn) mice separability of saline injected mice which were exposed to shock before getting environmental context (Shock Context). Both normal and trisomic mice receiving Shock Context treatment do not associate novel environments with adverse stimulus. Down Syndrome and normal mice were unseparable using PCA alone but are easily separable by cPCA at all values of the contrastive parameter $\lambda$.  The mice are also easily separable by UCA.  However, cPCA++ does not have similar separation performance compared to cPCA and UCA.}
  \label{fig:Mouse}
\end{figure*}
% \textbf{ I THINK YOUR ORIGINAL WRITEUP WAS BACKWARDS AND FLIPPED SHOCKED VS UNSHOCKED. ALSO, IS THE TOTAL SAMPLE SIZE OF 1080 CORRECT?}
% nope not flipped. and yes, total samples they say on the UCI database was 1080, and each sample could be treated as if they were independent even tho some measurements were repeated on the same mice. 

% p.3 1st paragraph doesn¿t match w/ Figure & paragraph 2. Make more clear when normal vs trisomy is background vs target. This might make it repetitive, but is necessary for clarity.

We applied UCA to mouse proteomics data, which were also used by \cite{Abid} to illustrate cPCA performance. The study measured levels of 76 proteins in 570 normal and 510 trisomic (Down Syndrome) mice receiving various combinations of learning therapies (Context Shock vs. Shock Context) and drug (memantine vs. saline) \cite{Ahmed, Higuera, Abid}. Normal mice exposed to Context Shock will first be exposed to novel context then expoed to shock, and learn to associate novel contexts with adverse stimulus; trisomic mice will not learn this association. Under the Shock Context treatment, mice are first exposed to shock than exposed to novel context, resulting in normal and trisomic mice not associating the environment with adverse stimulus. The goal of the experiment was to assess whether memantine improved learning ability in trisomic mice and to identify subsets of protein that may be involved in this process.

We first replicate the analysis by \cite{Abid}. We want to extract patterns of variation in protein levels that can help discriminate normal from trisomic mice. Our target dataset consisted of protein data from normal and trisomic mice which received Shock Context and were given saline. However, natural variation, arising from factors such as age and gender, may dominate this dataset and obscure the variation of interest due to trisomy. To remove this natural variation, we contrasted the target data with a background dataset consisting of normal Context Shocked mice who had been given saline. As natural variation is likely present in both the target and the background, patterns that explaining variation in the target but not the background may likely to related to trisomy.

Figure \ref{fig:Mouse} shows the data projected to the first two components identified by PCA, cPCA, cPCA++, and UCA. Normal and trisomic mice are not well-separated in the PCA results, showing that dominant variation in the target data indeed does not stem from trisomy. In contrast, the two mouse groups are much more clearly separated in the cPCA results. While this separation is noticeable for each of the tuning parameter values we tried, the actual projected data can vary considerably, and the optimal tuning parameter value remains unclear.
cPCA++, which does not require specifying tuning parameter here because the number of samples exceeds the number of features, shows better separation than PCA but does not perform as well as cPCA. UCA performs as well as cPCA but without requiring a tuning parameter.

\begin{figure*}[th!]
  \centering
  \includegraphics[width = 0.8\textwidth]{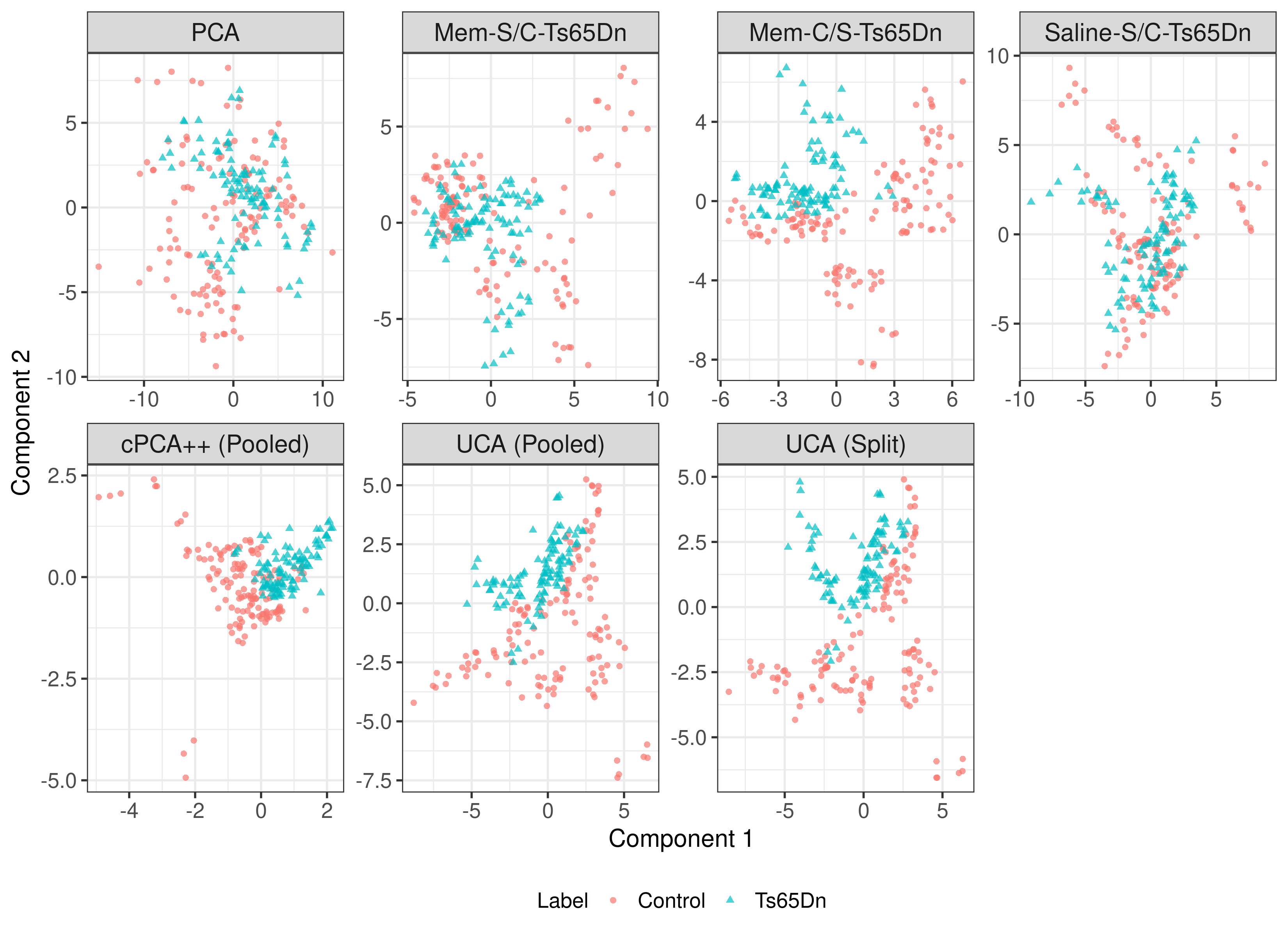}
  \caption{Mouse Protein Expression: Separability of normal and trisomic Context Shocked mice injected with saline. UCA with a background dataset of Shock Context trisomic mice injected with memantine (Mem-S/C-Ts65Dn), UCA with a background dataset of Context Shock trisomic mice injected with memantine (Mem-C/S-Ts65Dn), and UCA with a background dataset of Shock Context trisomic mice injected with saline (Saline-S/C-Ts65Dn). Pooled: all background datasets (Mem-S/C-Ts65Dn, Mem-C/S-Ts65Dn, Saline-S/C-Ts65Dn) are pooled into one background. Split: UCA using multiple individual backgrounds separately.}
  \label{fig:MouseSplitStack}
\end{figure*}

We next repeat the same analysis, but this time using Context Shocked mice given saline as our target dataset. \cite{Abid} did not consider this analysis, where separating normal and trisomic Context Shocked mice proves to be a much more challenging problem. Figure \ref{fig:MouseSplitStack} shows that normal and trisomic mice are again not well-separated by the first two components learned by PCA.
Here we extract patterns of variation unique to normal mice and so applied UCA using three trisomic mouse datasets as backgrounds.

The results show that a single background dataset alone cannot separate normal and trisomic mice well. Pooling the three individual backgrounds together achieves slightly better separability compared to some of the individual backgrounds. However, naively constrasting multiple backgrounds simultaneously without pooling allows our proposed UCA to remove variability specific to each background and results in the best separability.

\begin{figure*}[!ht]
    \begin{subfigure}[t]{\linewidth}
    \centering
    \includegraphics[width=\textwidth]{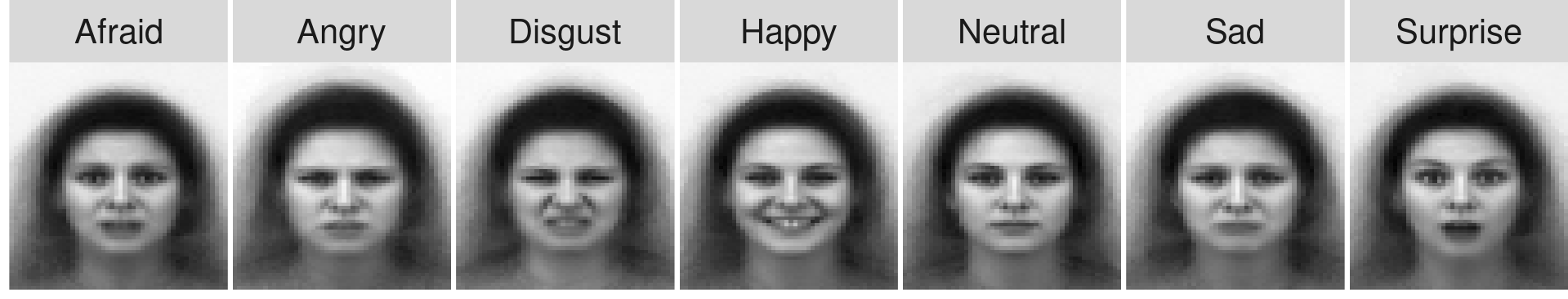}
    \caption{}
    \label{fig:faces-mean}
    %Average Faces for Each Emotion
    \end{subfigure}\\[1ex]
    \begin{subfigure}[b]{.49\linewidth}
    \centering
    \includegraphics[width=\textwidth]{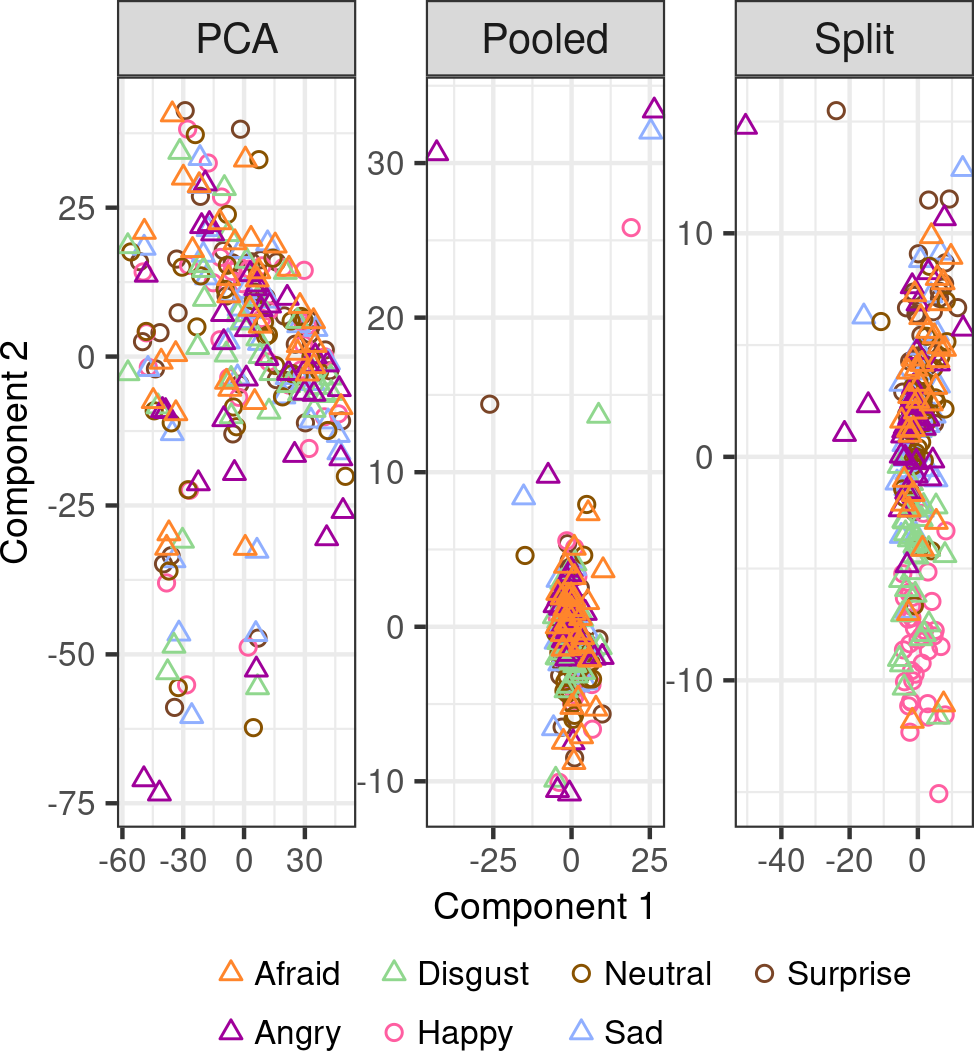}
    \caption{}
    \label{fig:faces-projected}
    \end{subfigure}
    \begin{subfigure}[b]{.49\linewidth}
    \centering
    \includegraphics[width=\textwidth]{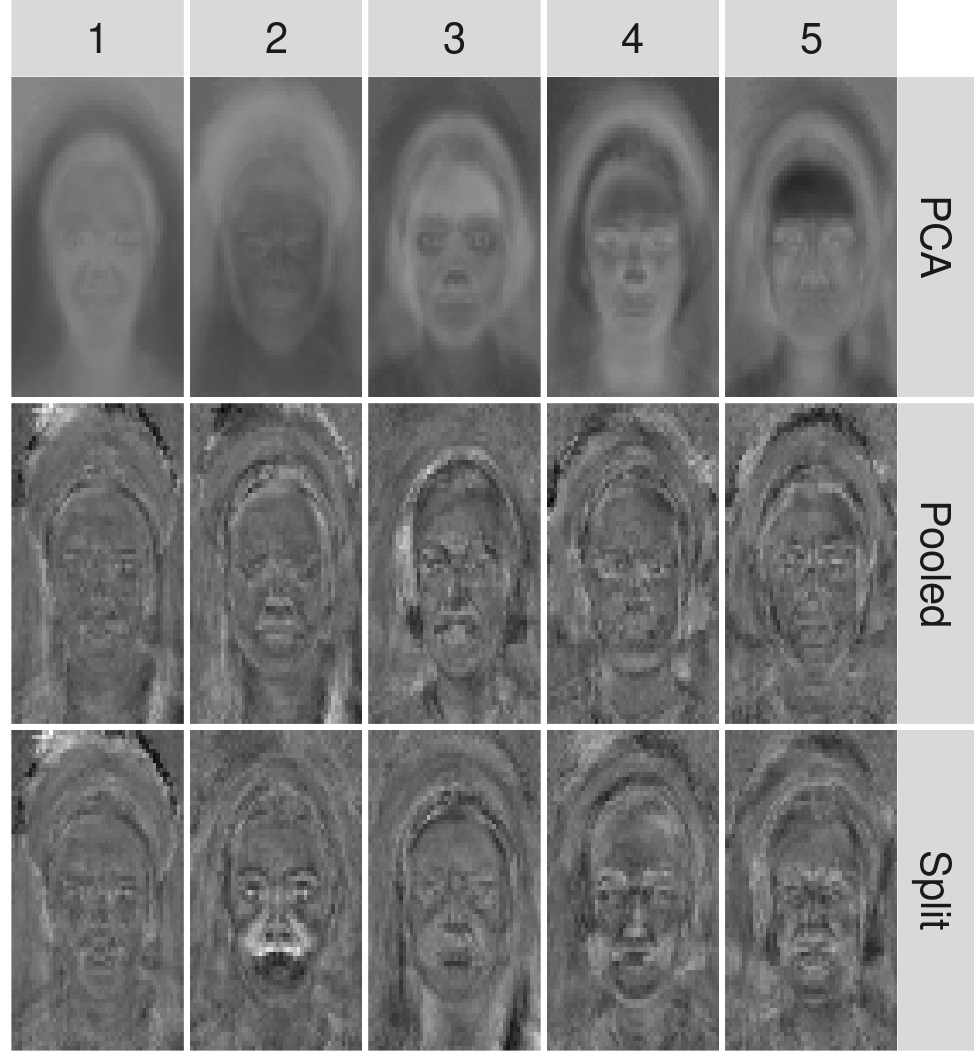}
    \caption{}
   \label{fig:faces-eigenfaces}
    \end{subfigure}
    \caption{Analysis of KDEF faces. (a) Each emotion averaged over all female actors. (b) Faces projected onto the first two components of PCA, UCA with a pooled background, and UCA using multiple backgrounds. Shapes distinguish negative (triangle) and non-negative (circle) emotions. (c) Top 5 eigenfaces of female emotions produced by PCA, UCA with a pooled background, and UCA using multiple backgrounds.}
\end{figure*}

\subsection{\label{sec:faces}Discovering eigenfaces of emotion}

        We further illustrate the advantages of contrasting multiple background datasets with UCA using the Karolinska Directed Emotional Faces (KDEF) by \cite{Calvo2008}, which captured images of seven emotions from five different angles from 70 amateur actors in either a practice or final session. Here, our target data includes all emotions from the final session. Figure \ref{fig:faces-mean} presents averaged images of each emotion calculated from all female actors in their final session. 

We focus on uncovering variation unique to the ``afraid'' emotion using all final session pictures from every emotion from female actors. Since this target contains six other emotions, standard PCA will not provide variation unique to ``afraid''. Instead, we leverage images from the practice session to serve as background data. We construct six separate backgrounds, corresponding to the six other emotions, and contrast out their variation using UCA. For comparison, we also pooled these images together into a single background dataset and performed UCA using the pooled background.

\begin{figure*}[th!]
  \centering
  \includegraphics[width = 0.9\textwidth]{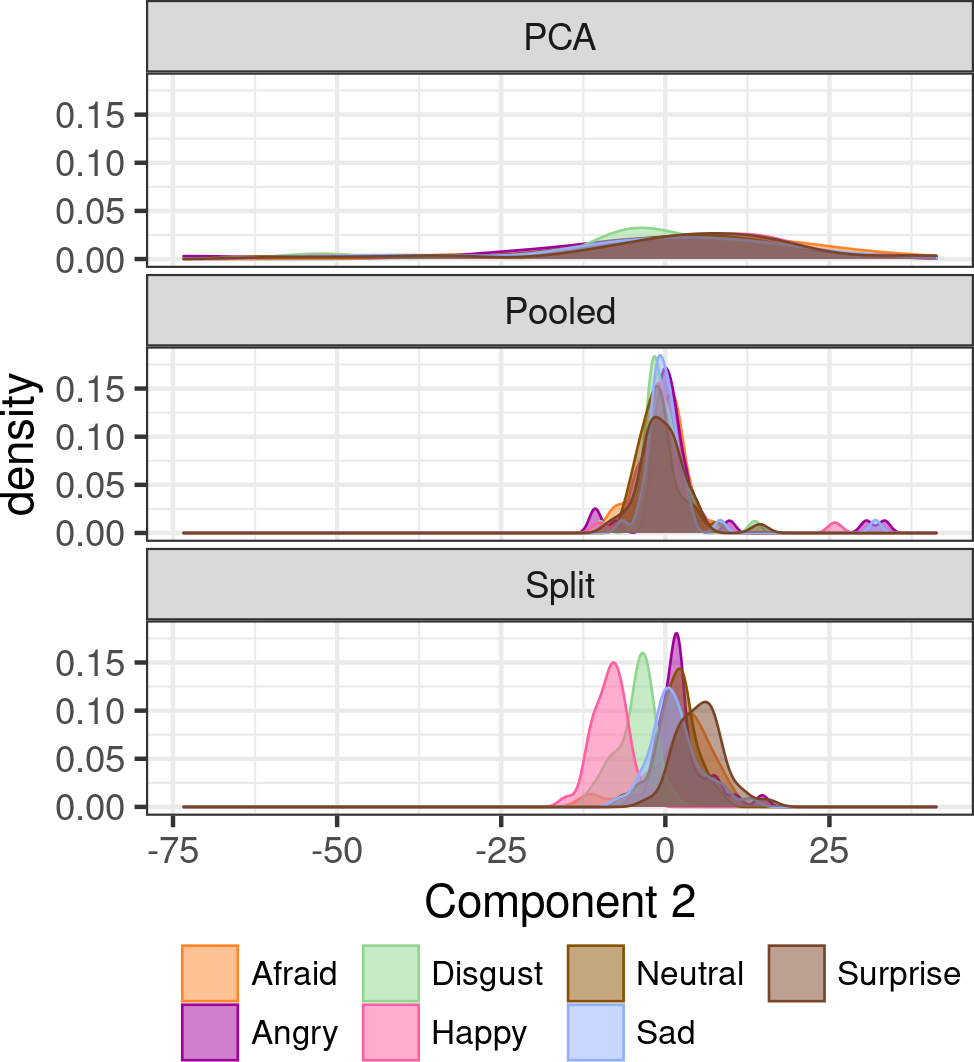}
  \caption{Density Plots, colored by emotions, of 2nd Component found by PCA, UCA with backgound of all emotional faces except for the afraid emotion pooled (Pooled), and UCA with background of all emotional faces except for the afraid emotion treated separately (Split). This is the same analysis in figure \ref{fig:faces-projected}, except using densities may help readers visualize the difference between pooling and splitting background data.}
  \label{fig:FacesDensity}
\end{figure*}

Figure \ref{fig:faces-projected} projects the target data onto the first two directions of variation calculated using each method. The emotions are entirely intermixed when projected onto components produced by PCA and by UCA using a single pooled background. In contrast, the emotions are much more separable when using UCA with multiple separate backgrounds. In particular, ``afraid'' faces are very separate from the ``disgust'' and ``happy'' faces. This indicates that UCA with multiple backgrounds can identify patterns of variation that can distinguish ``afraid'' from the other emotions. Supplemental figure \ref{fig:FacesDensity} plots the second component density to better show the differences in variability betwwen PCA, pooling, and splitting.

Figure \ref{fig:faces-eigenfaces} is a visualization of the top five directions of variation produced by each method as faces. These are called eigenfaces, and this visualization technique is useful for interpreting the top components as facial features \cite{turk1991eigenfaces}. PCA eigenfaces generally represent features common to all the emotions. UCA eigenfaces using a pooled background seem to highlight the eyebrows and upper lips, though it is difficult to discern. On the other hand, eigenfaces produced using UCA using multiple separate backgrounds highlight eyebrows, eyes, nostrils, and nasolabial folds. These are especially clear in the second eigenface and accord with intuition about which features would likely be most useful in distinguishing ``afraid'' from other emotions.

\subsection{\label{sec:product-svd} Product-SVD algorithm simulation}

\begin{figure}[!tpb]
    \centering
  \includegraphics[width = 0.5\textwidth]{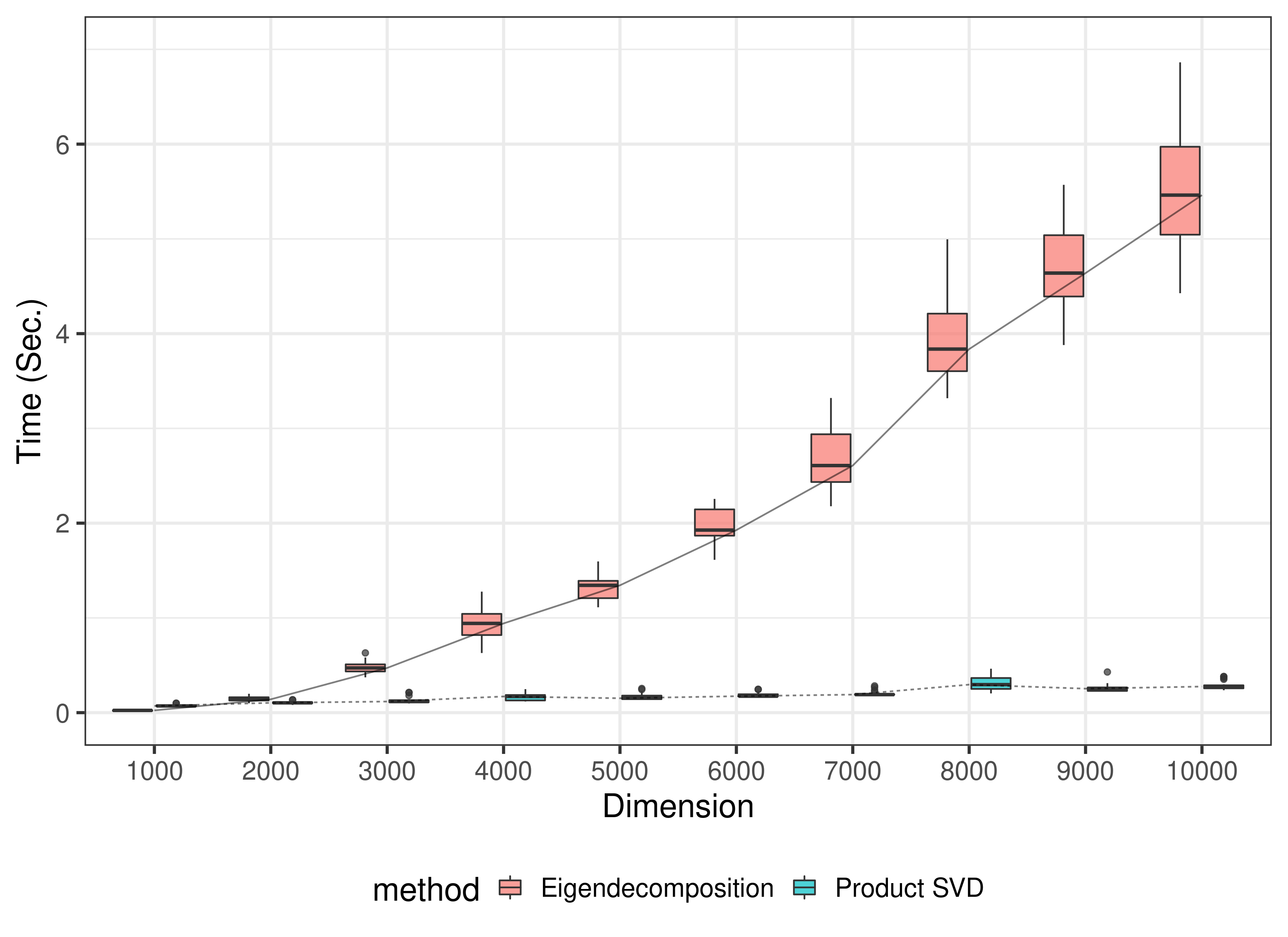}
  \caption{UCA and cPCA implementation using Product SVD vs. eigendecomposition for high-dimensional data: 25 random $100 \times p $ target and background matrices are generated from a standard normal distribution and where $p$, the dimension varied from 1,000 to 10,000 in steps of 1,000. Box plots of time (in seconds) is plotted for both eigendecomposition and the product SVD method. For small $p$, there is a negligible difference between the methods. However, as dimension $p$ increases, the Product SVD is significantly faster.}
  \label{fig:computational_perf}
\end{figure}

To demonstrate the speed improvements of the Product SVD method compared to the current fastest implementations of eigendecomposition in high-dimensions, we conduct a simulation study with 25 sample $100 \times p$ target and background data matrices generated from a standard normal distribution with $p$ varying from 1.000 to 10,000 in steps of 1,000. To ensure a fair comparison, we leverage C++ in both implementations using a custom RcppArmadillo (\cite{rcpparmadillo}) function for the Product SVD method and the RSpectra package (0.16-0) by \cite{Rspectra} for the eigendecomposition method. RSpectra is a package designed for large-scale eigendecompositions based off the C++ Spectra library. We use the microbenchmark package by \cite{microbenchmark} to ensure accurate timings. Our benchmark does not take into account the additional cost of forming the $p\times p$ covariance matrices, which would only exacerbate the difference between the two methods in real world applications.

Figure \ref{fig:computational_perf} shows box plots of time (in seconds) versus the dimension, $p$, colored by method, and summarizes the results of the simulation study. As the dimension $p$ increases, the computational time of our Product SVD method increases much slower than the current eigendecomposition implementations. It should be mentioned that for $p < 1000$ the Product SVD method is slower due to overhead of additional computation on small matrices. In general, for low-dimensional settings where $n \geq p$, the Product SVD will be negligibly slower because of the additional QR, SVD, matrix products, and sort computations. 
\section{Discussion}

In many data analytics settings, we are interested in removing uninteresting variation that contaminate the data of interest.
Here we proposed UCA, a tuning parameter free contrastive learning method that simplifies cPCA while also substantially improving it by accommodating multiple background datasets. We demonstrate UCA's usefulness and superiority in several examples. % To do this, we first solved the tuning parameter problem which plague current methods. UCA does not require user input in selecting the best contrastive parameter; rather it works by maximizing the target data variability subject to keeping the variability in each background small using weak duality of Lagrangians. 

UCA is computationally fast and easily extensible to high-dimensional data because it does not require constructing nor storing a large covariance matrices (see Methods). Further, no additional post-hoc clustering method is necessary to choose the appropriate tuning parameter.

Like cPCA, the choice of background(s) still plays a pivotal role in the directions found by UCA. If two background datasets contain highly correlated information, the algorithm will weight them accordingly. For example, in the event of identical backgrounds, only one background will be considered.

While the goal of UCA is not always to find separability between groups, UCA emphasizes finding variation in the target data scarcely seen in the background data. Therefore switching the roles of the background and target data may not result in the same target data separation. For example, if our target data consists of all facial emotions in section \ref{sec:faces}, and our background consisted of negative emotions (Afraid, Angry, Disgust, Sad), we hope to see less variability in negative emotions and preserve variability in non-negative emotions (Happy, Surprise, Neutral).
Conversely, using non-negative emotions as the background instead would preserve the variability in negative emotions, and decrease the variability in non-negative emotions. For maximal data seperation, we recommend using groups with larger variation as the background(s).

We have released the code for UCA as an R package, along with documentation and examples exhibited in this paper. 

\section{Acknowledgement}
Sandia National Laboratories is a multimission laboratory managed and operated by National Technology and Engineering Solutions of Sandia, LLC., a wholly owned subsidiary of Honeywell International, Inc., for the U.S. Department of Energy's National Nuclear Security Administration under contract DE-NA-0003525.

\bibliographystyle{jasa3}
\bibliography{mybib}

\section{Supplementary}
\end{document}